\documentclass[conference]{IEEEtran}

\usepackage{amsmath,amssymb,amsfonts}
\usepackage{graphicx}
\usepackage{textcomp}
\usepackage{cite}
\usepackage{algorithm}
\usepackage{algpseudocode}
\usepackage{booktabs} 
\usepackage{subfig}   
\begin{document}

\title{Detection of Anomalous Behavior in Robot Systems Based on Machine Learning}

\author{
    \IEEEauthorblockN{Mahfuzul I. Nissan}
    \IEEEauthorblockA{\textit{Department of Computer Science} \\
    \textit{University of New Orleans}\\
    New Orleans, LA, USA \\
    minissan@uno.edu}
    \and
    \IEEEauthorblockN{Sharmin Aktar}
    \IEEEauthorblockA{\textit{Department of Computer Science} \\
    \textit{St Mary's University}\\
    San Antonio, TX, USA \\
    saktar@stmarytx.edu}
}

\maketitle

\begin{abstract}
Ensuring the safe and reliable operation of robotic systems is paramount to prevent potential disasters and safeguard human well-being. Despite rigorous design and engineering practices, these systems can still experience malfunctions, leading to safety risks. In this study, we present a machine learning-based approach for detecting anomalies in system logs to enhance the safety and reliability of robotic systems. We collected logs from two distinct scenarios using CoppeliaSim and comparatively evaluated several machine learning models, including Logistic Regression (LR), Support Vector Machine (SVM), and an Autoencoder. Our system was evaluated in a quadcopter context (Context 1) and a Pioneer robot context (Context 2). Results showed that while LR demonstrated superior performance in Context 1, the Autoencoder model proved to be the most effective in Context 2. This highlights that the optimal model choice is context-dependent, likely due to the varying complexity of anomalies across different robotic platforms. This research underscores the value of a comparative approach and demonstrates the particular strengths of autoencoders for detecting complex anomalies in robotic systems.
\end{abstract}

\begin{IEEEkeywords}
Robotic Systems, Anomaly Detection, Machine Learning, System Logs, Logistic Regression, Autoencoder, Contextual Evaluation, Coppeliasim, Quadcopter, Pioneer Robot
\end{IEEEkeywords}

\section{Introduction}
The increasing integration of robotic systems into industrial, commercial, and domestic environments has led to significant advancements in automation and efficiency. However, this widespread adoption brings a critical challenge: ensuring the safety and reliability of these complex systems. Malfunctions or anomalous behaviors can lead to operational failures, equipment damage, and pose significant risks to human safety. Consequently, the development of robust methods for detecting and preventing such anomalies is a paramount concern in modern robotics.

Traditional approaches to robot safety often rely on reactive, rule-based systems that trigger alerts only after a deviation has already occurred. These methods are often insufficient for preventing failures, as the delay between an anomaly's onset and its detection can be catastrophic. To address this limitation, this research explores a proactive, machine learning-based approach for anomaly detection in robotic systems.

In this paper, we present an anomaly detection system that analyzes robot operational logs from a simulated environment. We leverage both supervised machine learning models, specifically Logistic Regression and Support Vector Machines and an unsupervised autoencoder to identify abnormal behavior. Our primary contribution is a comparative analysis of these methods across two distinct robotic contexts: quadcopter navigation and the coordinated movement of Pioneer robots. Our findings demonstrate the effectiveness of these techniques and highlight how the optimal model choice is dependent on the specific operational context.

The remainder of this paper is organized as follows: Section II reviews related work in robot anomaly detection. Section III provides background on proactive detection and the motivation for our approach. Section IV details our methodology. Section V presents the experimental setup, data collection, and results. Finally, Section VI concludes the paper and discusses directions for future work.

\section{Related Work}

The challenge of ensuring robot safety has spurred a shift from traditional rule-based systems to proactive, learning-based anomaly detection. A primary focus in this area is the analysis of system-level data, including operational logs and sensor readings, to predict failures before they occur. In the specific domain of robot navigation, Ji et al. \cite{ji2022proactive} proposed a proactive anomaly detection network (PAAD) that uses multi-sensor fusion to predict the probability of future failure. This highlights a growing trend of applying sophisticated data analysis techniques to enhance the reliability of autonomous systems.

Deep learning has emerged as a particularly powerful tool for automating anomaly detection from complex system data. Among these techniques, autoencoders have been widely explored for their ability to learn baseline patterns of normal behavior in an unsupervised manner. For instance, Olivato et al. \cite{olivato2019comparative} conducted a comparative analysis of autoencoders for detecting cyber-security attacks by transforming system logs into images. Similarly, Park et al. \cite{park2018multimodal} proposed an LSTM-based variational autoencoder (LSTM-VAE) to fuse multimodal sensory signals for detecting anomalies in robot-assisted feeding, while Utkin et al. \cite{utkin2016detection} utilized an autoencoder for dimensionality reduction to establish a detection threshold. These studies collectively demonstrate the versatility and effectiveness of autoencoders in identifying subtle deviations from normal operational patterns.

A unifying principle across robotics and broader cyber-physical systems research is the analysis of direct, low-level system artifacts to uncover ground-truth activity. In robotics, this often translates to using raw telemetry or system logs, while in other domains it has involved memory snapshots, byte-level storage, and network traffic. Our previous work has shown that user activity can be reconstructed by carving artifacts directly from memory snapshots and byte-level storage in both relational \cite{nissan2023301503},\cite{nissan2022forensic},\cite{nissan2025memtracedbreconstructingmysqluser} and NoSQL systems \cite{NISSAN2025301929}. This methodological focus on raw data analysis is motivated by the inherent limitations of system audit logs, which can be altered or bypassed. We have also demonstrated that deep learning models, particularly contractive autoencoders and their ensembles are highly effective at detecting sophisticated network threats by learning baseline traffic patterns from audit data \cite{AKTAR2023103251}, \cite{aktar2022intrusion}, \cite{aktar2024syscon}, \cite{aktar2024advancing}. This principle of trusting direct system artifacts over potentially fallible high-level logs is directly transferable to robotics, where operational logs are the primary evidence of system behavior. Although these studies address different application areas, they collectively reinforce the value of direct, low-level data analysis as a foundation for reliable anomaly detection.

This current research builds on that principle by applying direct system-level data analysis to behavioral classification in robotics. While prior work highlights the strengths of advanced architectures such as LSTM-VAE or ensemble autoencoders, there remains a lack of comparative studies that evaluate classic supervised models alongside unsupervised autoencoder approaches on the same set of operational logs. Our work addresses this gap by systematically evaluating these distinct machine learning methodologies within a controlled, simulated robotic environment.

\section{Background and Motivation}
The task of ensuring robot safety can be broadly categorized into two paradigms: reactive and proactive detection. Reactive systems, which form the basis of many traditional safety protocols, are designed to respond to failures after they have happened. They typically rely on predefined thresholds and explicit rules (e.g., if motor torque exceeds value X, halt operation). While essential, these methods are fundamentally limited; they cannot anticipate novel failures and their response time may be too slow to prevent significant damage or harm.

The limitations of reactive approaches create a strong motivation for proactive anomaly detection, which aims to identify the precursors to failure before a critical event occurs. Proactive systems move beyond static rules by learning the complex patterns inherent in a robot's normal operational data. By building a model of what constitutes "normal," these systems can flag subtle deviations that may indicate an impending malfunction. Machine learning, and deep learning in particular, is exceptionally well-suited for this task due to its ability to automatically learn intricate, high-dimensional patterns from sensor and system log data.

Among various machine learning techniques, autoencoders have emerged as a particularly promising tool for this purpose. An autoencoder is a neural network trained to reconstruct its input. By training it exclusively on data from normal operations, the network becomes highly adept at this reconstruction task. When it encounters anomalous data, which deviates from the patterns it has learned, its reconstruction error increases significantly. This principle allows the autoencoder to function as a powerful, unsupervised anomaly detector, capable of identifying unexpected system behaviors without prior knowledge of specific failure modes. This research is motivated by the need for such advanced, proactive systems to enhance the security and reliability of modern robotic platforms.

\section{Methodology}
In this section, we describe the methodology used in our work on two scenarios simulated in the CoppeliaSim environment. The first scenario, referred to as Context 1, involves the navigation of a quadcopter. The second scenario, referred to as Context 2, involves the navigation of a Pioneer robot. Our approach involved following the $D^{*}$ path planning algorithm to create normal and anomalous data and detecting anomalies using machine learning models. The methodology section is divided into three subsections. The first subsection provides general information about the $D^{*}$ algorithm, which was used for path planning. The second and third subsections provide general information about the use of Support Vector Machine (SVM) and Logistic Regression (LR) machine learning models, respectively, for anomaly detection.

\subsection{$D^{*}$ Algorithm}
The $D^{*}$ algorithm, also known as D-Star, is a popular path planning algorithm used in robotics and autonomous systems. It is designed to handle changing environments by enabling efficient and dynamic re-planning. The algorithm operates on a grid-based map representation of the environment, where each cell stores information about the cost of reaching that cell and an estimate of the remaining cost to the goal.

During exploration, the algorithm selects the cell with the minimum cost and updates the costs of its neighboring cells based on factors such as distance and the presence of obstacles. It dynamically adjusts the costs as it encounters changes in the environment, such as the addition or removal of obstacles. The $D^{*}$ algorithm continues to explore and update the costs until it reaches the goal or determines that no feasible path exists. It provides a dynamic path from the start location to the goal, adapting to changes in the environment as they occur. Algorithm \ref{alg:dstar} shows the $D^{*}$ algorithm.

\begin{algorithm}
\caption{$D^{*}$ Algorithm}
\label{alg:dstar}
\begin{algorithmic}[1]
    \State \textbf{Input:} Start cell, Goal cell
    \State \textbf{Output:} Optimal path from start to goal
    \State Initialize costs and estimates of all cells except goal;
    \State Set cost of goal cell to 0;
    \State Initialize an empty Open List;
    \While{Open List is not empty}
        \State Select cell with minimum cost from Open List;
        \If{Goal cell reached}
            \State \textbf{return} optimal path;
        \Else
            \State Update costs and estimates of neighboring cells;
            \If{Changes detected in the environment}
                \State Update costs and estimates accordingly;
            \EndIf
        \EndIf
    \EndWhile
    \State \textbf{return} dynamic path from start to goal;
\end{algorithmic}
\end{algorithm}

\subsection{Logistic Regression}
Logistic regression is a widely used statistical model for binary classification tasks, where the goal is to predict the probability of an event occurring or not occurring based on a set of input features. It is a linear classifier that maps the input features to the probability of the binary outcome using a logistic function.

The logistic regression algorithm calculates a weighted sum of the input features, applying these weights to determine their contribution to the final prediction. The logistic function, also known as the sigmoid function, is then applied to this sum to transform it into a value between 0 and 1. This value represents the predicted probability of the event occurring.

During the training phase, the model adjusts the weights of the input features to maximize the likelihood of the observed data. This is typically done using optimization algorithms such as gradient descent. Once trained, the logistic regression model can make predictions by applying the learned weights to new instances and transforming the weighted sum using the logistic function.

Equation for logistic regression:
\begin{equation}
P(y=1|x) = \frac{1}{1 + e^{-(\beta_0 + \beta_1x_1 + \beta_2x_2 + \dots + \beta_nx_n)}}
\end{equation}
Here, $P(y=1|x)$ represents the probability of the event (e.g., class label y being 1) given the input features $x$. The coefficients $\beta_0, \beta_1, \beta_2, \dots, \beta_n$ correspond to the weights assigned to each feature $x_1, x_2, \dots, x_n$. The logistic function is $\frac{1}{1+e^{-z}}$, used to transform the weighted sum $z$ into a probability value between 0 and 1.

\subsection{Support Vector Machine (SVM)}
Support Vector Machine (SVM) is a powerful supervised machine learning algorithm used for classification and regression tasks \cite{huang2018applications}. It works by finding an optimal hyperplane that separates the data points into different classes, maximizing the margin between the classes. In SVM, the objective is to find the best decision boundary that maximizes the margin, which is the distance between the decision boundary and the closest data points from each class, known as support vectors. These support vectors play a crucial role in defining the decision boundary and determining the overall performance of the SVM.

Mathematically, SVM aims to solve the following optimization problem:
\begin{equation}
\min_{w,b} \left( \frac{1}{2} ||w||^2 + C \sum_{i=1}^{N} \xi_i \right)
\end{equation}
constrained by:
\begin{equation}
y_i(w \cdot x_i + b) \ge 1 - \xi_i, \quad \xi_i \ge 0 \quad \forall i=1,2,\dots,N
\end{equation}
where $x_i$ is the feature vector of the i-th data point, $y_i$ is its corresponding class label (-1 or 1), $w$ is the weight vector, $b$ is the bias term, $\xi_i$ is the slack variable that allows for misclassifications, and C is a regularization parameter that controls the trade-off between maximizing the margin and minimizing the classification errors.

By solving this optimization problem, SVM finds the optimal values of $w$ and $b$ that define the decision boundary. The decision function can be expressed as:
\begin{equation}
f(x) = \text{sign}(w \cdot x + b)
\end{equation}
where $x$ is the input feature vector, and the sign function determines the predicted class label based on the position of the input vector with respect to the decision boundary.

In our project, we employed the linear kernel, a fundamental component of the SVM algorithm. The linear kernel is represented mathematically as:
\begin{equation}
K(x_i, x_j) = x_i \cdot x_j
\end{equation}
where $x_i$ and $x_j$ denote the input feature vectors. The linear kernel computes the dot product between these feature vectors, resulting in a linear function that captures the relationships between the data points.
\section{Experiments}
In this section, we present the experimental setup and implementation of the $D^{*}$ algorithm in CoppeliaSim to simulate the navigation of both a quadcopter and a Pioneer robot. The objective of these experiments was to implement and demonstrate the functionality of the $D^{*}$ algorithm in the CoppeliaSim environment for controlling the quadcopter and Pioneer robot. The focus was on successfully integrating and configuring the $D^{*}$ algorithm with both robot platforms, allowing them to navigate from their respective start locations to their goal locations within the defined workspace. While the $D^{*}$ algorithm itself was not modified, necessary adjustments were made to scale it to the CoppeliaSim workspace and ensure compatibility with both the quadcopter and Pioneer robot platforms.

\subsection{CoppeliaSim}
CoppeliaSim \cite{rohmer2013coppeliasim}, also known as V-REP (Virtual Robot Experimentation Platform), is a versatile and widely-used robotics simulation software. It provides a comprehensive platform for simulating, programming, and testing various robotic systems and algorithms. CoppeliaSim allows users to create virtual environments and models, interact with them in real-time, and simulate the behavior and performance of robotic systems. At its core, CoppeliaSim operates on a physics-based simulation engine, which accurately models the dynamics and interactions of objects within the simulated environment. This enables realistic simulations of robot movements, sensor readings, and interactions with the virtual world. CoppeliaSim supports a wide range of robot models, sensors, and actuators, allowing users to simulate diverse robotic applications.

\subsection{Implementation of Quadcopter in CoppeliaSim}
First, we created a workspace in CoppeliaSim with the specifications $W = [xmin = -2.5, xmax = 2.5, ymin = -2.5, ymax = 2.5]$, representing the default floor dimensions. This workspace served as the environment in which the quadcopter would navigate. Next, we transformed the coordinates of the obstacles to fit within the CoppeliaSim workspace. This step ensured that the obstacles were correctly positioned and aligned with the virtual environment. To define the start and goal locations, we created two cuboid objects in CoppeliaSim. The start location represented the initial position of the quadcopter, while the goal location indicated the desired destination. Using the $D^{*}$ algorithm, we computed a path connecting the start and goal locations within the workspace. The algorithm considered the presence of obstacles and dynamically updated the path based on any changes in the environment. Additionally, we introduced a safety parameter to ensure that the quadcopter maintained a safe distance from the workspace boundaries. This parameter prevented the quadcopter from approaching the edges of the workspace, reducing the risk of collisions and ensuring safe navigation. Finally, we implemented the quadcopter model in CoppeliaSim and programmed it to follow the computed path from the start to the goal location. The quadcopter's movement was synchronized with the $D^{*}$ algorithm, allowing it to traverse the workspace while avoiding obstacles and maintaining a safe distance from the boundaries.

\subsection{Implementation of Pioneer Robot in CoppeliaSim}
We implemented two Pioneer P3DX robots in the CoppeliaSim environment as part of our experimental setup. These robots were initialized at random states, with one robot positioned at the starting location and the other robot placed at the goal location. The goal was to exchange their initial positions while primarily following the $D^{*}$ path. To achieve this, we generated a collision-free path that enabled the robots to navigate through the workspace and exchange their positions while adhering to the computed $D^{*}$ path. These paths were carefully designed to ensure the robots could move safely without colliding with obstacles or each other. Similar to the quadcopter implementation, we utilized the $D^{*}$ algorithm to compute the paths connecting the start and goal locations for the Pioneer P3DX robots. The algorithm accounted for obstacles and dynamically adjusted the paths as the robots traversed the environment. During the experiment, the robots followed their respective paths, executing the position exchange while adhering to the computed paths. The movement of the robots was synchronized with the $D^{*}$ algorithm, allowing them to successfully navigate through the workspace and exchange their initial positions.

\begin{figure*}[!t]
    \centering
    \includegraphics[width=.8\textwidth]{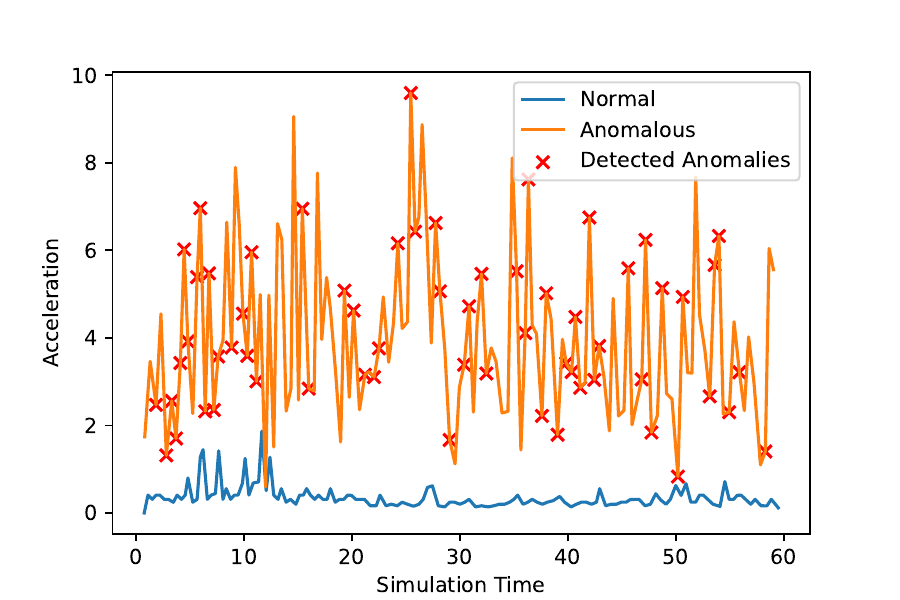}
    \caption{Acceleration vs Simulation Time for Context 1}
    \label{fig:accel_context1}
\end{figure*}

\begin{figure*}[!t]
    \centering
    \includegraphics[width=.8\textwidth]{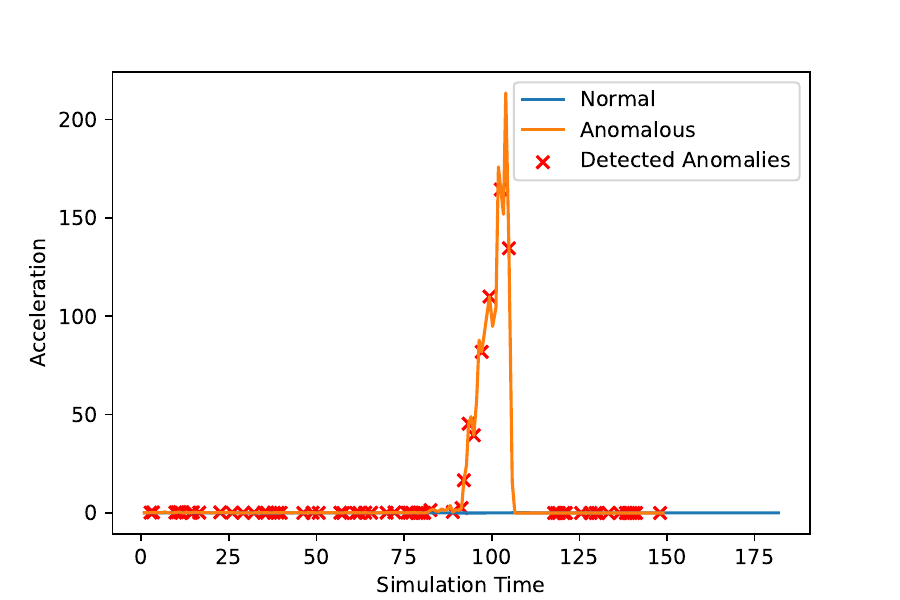}
    \caption{Acceleration vs Simulation Time for Context 2}
    \label{fig:accel_context2}
\end{figure*}

\begin{figure*}[!t]
    \centering
    \includegraphics[width=.8\textwidth]{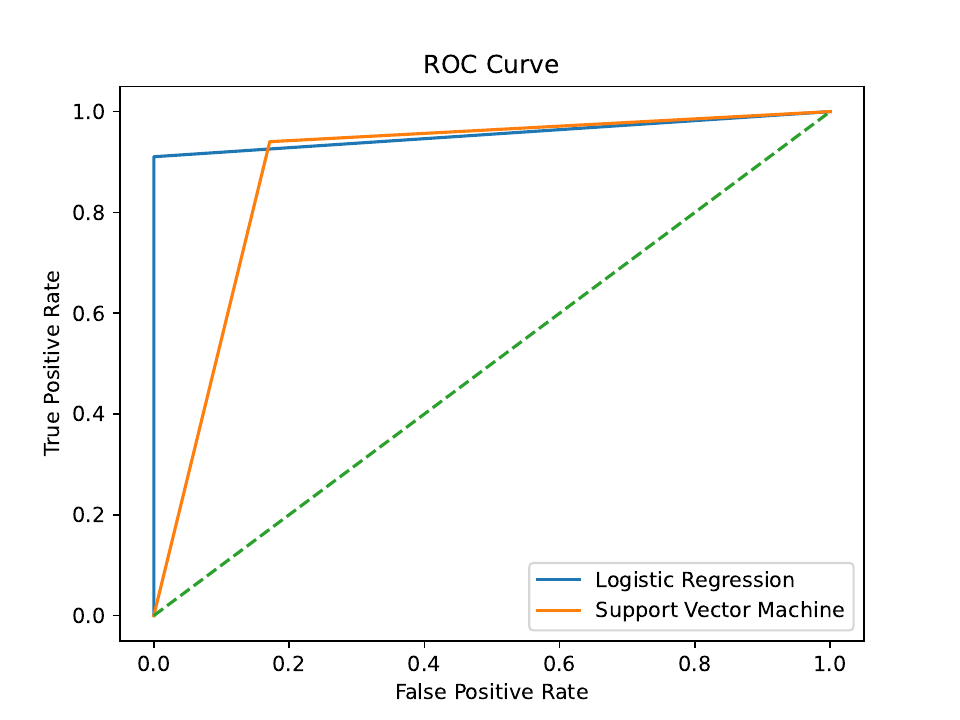}
    \caption{ROC Curves for Context 1}
    \label{fig:roc_context1}
\end{figure*}

\begin{figure*}[!t]
    \centering
    \includegraphics[width=.8\textwidth]{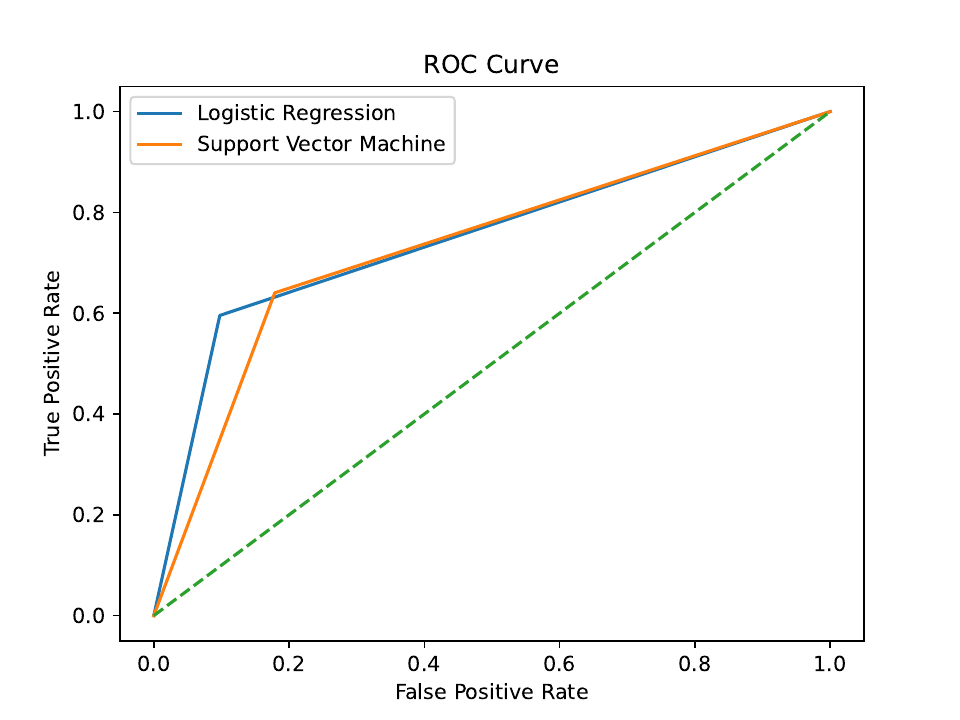}
    \caption{ROC Curves for Context 2}
    \label{fig:roc_context2}
\end{figure*}

\subsection{Data Collection}
To gather comprehensive data on the behavior and dynamics of both the quadcopter and Pioneer robot, we conducted meticulous data collection during simulation runs. Two scenarios, namely the normal and abnormal scenarios, were employed to simulate different operational conditions.

In the normal scenario, both the quadcopter and Pioneer robot exhibited smooth and gradual movements within the simulated environment. They followed predefined paths and maintained stable position, orientation, velocity, acceleration, and angular velocity throughout the simulation. This scenario aimed to capture the baseline behavior of both robots.

In contrast, the anomalous scenario introduced sudden and unpredictable movements to mimic unexpected events or disturbances. Random position offsets and joint velocity fluctuations were applied to both the quadcopter and Pioneer robot, causing them to deviate from their intended paths. This scenario aimed to simulate challenging situations and assess the robustness of their control systems.

During each simulation run, we logged various data points for both the quadcopter and Pioneer robot, including the timestamp, position, orientation, velocity, acceleration, and angular velocity. This information was recorded at regular intervals throughout the simulation to capture the dynamic behavior of both robots. All the logged data for the quadcopter and Pioneer robot was saved to separate text files for further analysis and processing.

The acceleration profiles of the quadcopter and Pioneer robot were analyzed to better understand their behavior and dynamics in different scenarios. Figures \ref{fig:accel_context1} and \ref{fig:accel_context2} illustrate the relationship between acceleration and time for both normal and anomalous logs in Context 1 and Context 2, respectively. In Context 1, the normal scenario (Figure \ref{fig:accel_context1}) shows smooth and consistent acceleration for both robots throughout the simulation, reflecting stable and predictable movements within the simulated environment. In contrast, the anomalous scenario (Figure \ref{fig:accel_context1}) introduces sudden changes in acceleration for both robots, indicating the presence of unexpected events or disturbances. Notably, our anomaly detection system was able to correctly identify these points as anomalous.

Similarly, in Context 2, the acceleration profiles for normal and anomalous logs are shown in Figure \ref{fig:accel_context2}. The normal scenario (Figure \ref{fig:accel_context2}) displays consistent acceleration patterns, while the anomalous scenario (Figure \ref{fig:accel_context2}) exhibits abrupt changes in acceleration, indicating challenging situations that affect the robots' movements. These acceleration profiles offer valuable insights into the dynamic behavior of the quadcopter and Pioneer robot under various operational conditions. They highlight the differences between normal and anomalous scenarios and allow for a thorough analysis of the control systems' robustness.

\begin{table*}[!t]
\caption{Comparison of Classification Results for Context 1 (Average of 10 iterations)}
\label{tab:context1_results}
\centering
\begin{tabular}{@{}lccccc@{}}
\toprule
\textbf{Model} & \textbf{ROC Score} & \textbf{Precision} & \textbf{Recall} & \textbf{Accuracy} & \textbf{F1-score} \\ \midrule
LR    & 0.9552 & 0.9597 & 0.9562 & 0.9562 & 0.9561 \\
SVM   & 0.8844 & 0.8888 & 0.8832 & 0.8832 & 0.8830 \\ \bottomrule
\end{tabular}
\end{table*}

\begin{table*}[!t]
\caption{Comparison of Classification Results for Context 2 (Average of 10 iterations)}
\label{tab:context2_results}
\centering
\begin{tabular}{@{}lccccc@{}}
\toprule
\textbf{Model} & \textbf{ROC Score} & \textbf{Precision} & \textbf{Recall} & \textbf{Accuracy} & \textbf{F1-score} \\ \midrule
Autoencoder & 0.7675 & 0.8488 & 0.8038 & 0.8038 & 0.7888 \\
SVM         & 0.7308 & 0.7435 & 0.7453 & 0.7453 & 0.7427 \\
LR          & 0.7490 & 0.7804 & 0.7736 & 0.7736 & 0.7660 \\ \bottomrule
\end{tabular}
\end{table*}

\subsection{Results}
In this section, we present the results of our classification experiments for Context 1 and Context 2. The performance of two different models, Logistic Regression (LR) and Support Vector Machine (SVM), was evaluated using various metrics, including ROC score, precision, recall, accuracy, and F1-score. Each reported result represents the average of 10 iterations to ensure robustness and reliability of the findings.

For Context 1, the LR model demonstrated exceptional performance across all evaluation metrics. It achieved an impressive ROC score of 0.9552 and high values for precision (0.9597), recall (0.9562), accuracy (0.9562), and F1-score (0.9561) (see Table \ref{tab:context1_results}). These results suggest that LR effectively classified the data instances in Context 1. In contrast, the SVM model exhibited slightly lower performance in Context 1. It achieved an ROC score of 0.8844 and lower values for precision (0.8888), recall (0.8832), accuracy (0.8832), and F1-score (0.8830) compared to LR. The strong performance of Logistic Regression in this context suggests that the anomalies within the quadcopter's navigation logs were likely linearly separable. The introduced disturbances may have created clear, distinct shifts in the feature space that a simple linear model could easily capture.

Moving to Context 2, we introduced an additional model, the Autoencoder. In this context, the Autoencoder model achieved the best results among the three models. It obtained the highest ROC score of 0.7675 and demonstrated strong performance across all evaluation metrics, with a precision of 0.8488, recall of 0.8038, accuracy of 0.8038, and an F1-score of 0.7888. The Logistic Regression (LR) model also performed well, achieving a ROC score of 0.7490 and demonstrating good performance with precision, recall, accuracy, and F1-score values of 0.7804, 0.7736, 0.7736, and 0.7660, respectively. The SVM model achieved a ROC score of 0.7308, indicating a relatively lower discriminative ability compared to the other models (see Table \ref{tab:context2_results}). These findings suggest that the Autoencoder model is a highly effective and favorable approach for this context. The superior performance of the Autoencoder here indicates that the anomalies in the Pioneer robot's logs may have been more subtle or non-linear in nature. The Autoencoder's ability to learn a robust representation of normal behavior allowed it to flag deviations that the linear boundaries of SVM and LR found more challenging to isolate.

To provide a visual representation of the model performance, Figure \ref{fig:roc_context1} presents the ROC curve for Context 1, showcasing the performance of the LR and SVM models. The plot demonstrates the true positive rate (sensitivity) against the false positive rate (1 - specificity) at various classification thresholds. The ROC curve allows us to compare and assess the discriminative ability of the different models. Similarly, Figure \ref{fig:roc_context2} presents the ROC curve for Context 2, featuring the LR, SVM, and Autoencoder models. This curve illustrates the performance of the models in terms of their true positive rate and false positive rate at different classification thresholds.

\section{Conclusion and Future Work}
In conclusion, our study involved collecting system logs from different scenarios using Coppeliasim and developing a machine learning-based anomaly detection system. We evaluated the performance of this system in two contexts using different models, including Logistic Regression (LR), SVM, and an Autoencoder. Our results showed that the LR model demonstrated superior performance in Context 1, while the Autoencoder model proved to be the most effective in Context 2. This highlights that the optimal model choice can be context-dependent, likely due to the varying nature and complexity of anomalies across different robotic platforms and tasks.

For future work, we plan to explore sequence-based models like Long Short-Term Memory (LSTM) networks. Since system logs are inherently time-series data, an LSTM could capture temporal dependencies that our current models ignore, potentially allowing for even earlier and more accurate anomaly prediction. Furthermore, we intend to expand our evaluation to include scenarios with multiple, simultaneous anomalies to test the robustness of our system in more complex operational environments. Finally, we aim to deploy the most promising models onto physical hardware to validate their performance and real-time feasibility outside of a simulated environment.

\bibliographystyle{unsrt}
\bibliography{references}

\end{document}